\documentclass[conference]{IEEEtran}
\IEEEoverridecommandlockouts
\usepackage{subfigure}
\usepackage{cite}
\usepackage{amsmath,amssymb,amsfonts}
\DeclareMathOperator*{\argmax}{argmax}
\usepackage{graphicx}
\usepackage{textcomp}
\usepackage{xcolor}
\usepackage{algpseudocode}
\usepackage{algorithm}
\usepackage{float}
\usepackage{tabularx}
\usepackage{caption}

\def\BibTeX{{\rm B\kern-.05em{\sc i\kern-.025em b}\kern-.08em
    T\kern-.1667em\lower.7ex\hbox{E}\kern-.125emX}}
\begin{document}

\title{Local Navigation and Docking of an Autonomous Robot Mower using Reinforcement Learning and Computer Vision
}

\author{\IEEEauthorblockN{Ali Taghibakhshi}
\IEEEauthorblockA{\textit{Department of Mechanical Science} \\
\textit{and Engineering}\\
\textit{The University of Illinois}\\
\textit{at Urbana-Champaign}\\
Urbana, Illinois, USA \\
alit2@illinois.edu}
\and
\IEEEauthorblockN{Nathan Ogden}
\IEEEauthorblockA{\textit{John Deere} \\
\textit{Technology Innovation Center}\\
Champaign, Illinois, USA  \\
ogdennatahn@johndeere.com}
\and
\IEEEauthorblockN{Matthew West}
\IEEEauthorblockA{\textit{Department of Mechanical Science} \\
\textit{and Engineering}\\
\textit{The University of Illinois}\\
\textit{at Urbana-Champaign}\\
Urbana, Illinois, USA \\
mwest@illinois.edu}
}

\maketitle

\begin{abstract}
We demonstrate a successful navigation and docking control system for the John Deere Tango autonomous mower, using only a single camera as the input. This vision-only system is of interest because it is inexpensive, simple for production, and requires no external sensing. This is in contrast to existing systems that rely on integrated position sensors and global positioning system (GPS) technologies. To produce our system we combined a state-of-the-art object detection architecture, You Only Look Once (YOLO), with a reinforcement learning (RL) architecture, Double Deep Q-Networks (Double DQN). The object detection network identifies features on the mower and passes its output to the RL network, providing it with a low-dimensional representation that enables rapid and robust training. Finally, the RL network learns how to navigate the machine to the desired spot in a custom simulation environment. When tested on mower hardware, the system is able to dock with centimeter-level accuracy from arbitrary initial locations and orientations.
\end{abstract}


\section{Introduction}
The ever-growing field of autonomous vehicles is a research hotbed for applying Artificial Intelligence (AI) and Machine Learning (ML). In the past few years, there have been a wide variety of improvement in autonomous vehicles. These improvements include, but are not limited to, single agent scale tasks such as path planning, lane changing, and self driving, and multi-agent scale tasks such as collision avoidance and lane management. For instance, in \cite{cao2017optimal}, authors have investigated an optimal lane change decision model for autonomous vehicles in high-capacity urban roads. In recent studies \cite{karimi2019optimal, shahri2020optimal}, researchers have introduced a novel lane management in heterogeneous framework with autonomous vehicles.

Recently, many companies have invested efforts in developing ML-aided driving for increased comfort and safety of their vehicles. However, many autonomous driving and navigation systems still rely on sensing modalities such as Laser Range Finder (LRF), Light Detection and Ranging (LIDAR), and GPS, to name but a handful. These systems can be expensive and may induce further complications in the design of an autonomous vehicle. It is thus desirable to produce systems which use only vision as the control input.

Over the past decade, reinforcement learning techniques and algorithms have been able to solve complicated decision making problems, both in single-agent \cite{silver2016mastering, mnih2015human} and multi-agent frameworks \cite{vinyals2019grandmaster,shojaeighadikolaei2020demand, 9302981}.  With the rise of deep reinforcement learning, there have been multiple studies on implementing value-based learning methods, mostly deep Q-learning \cite{mnih2015human}, in the field of autonomous driving. Using a visual encoding, authors in \cite{chen2019model} utilized a low dimensional representation of the driving environment as the observation input for model-free RL agents, whose aim is to achieve urban autonomous driving.  \cite{sallab2017deep} has proposed a deep RL framework for autonomous driving, where the RL agents observe raw sensor outputs and take driving actions.

This paper focus on the docking procedure for the John Deere Tango robot mower, which is not reliable in it's current form. This mower was designed to dock using a guide loop wire buried underground, which leads the mower toward the charging station by inducing an electric current in the wire, which is then sensed by a detection system in the mower. Docking failures can occur when the wire is misplaced under the ground or there is a bump in the mower's path. This motivated us to investigate a system that it is more robust to variable initial position and environmental conditions.

\begin{figure}[ht]
\centering
\includegraphics[clip, trim=2.5cm 14cm 0.5cm 2.5cm, width=0.55\textwidth]{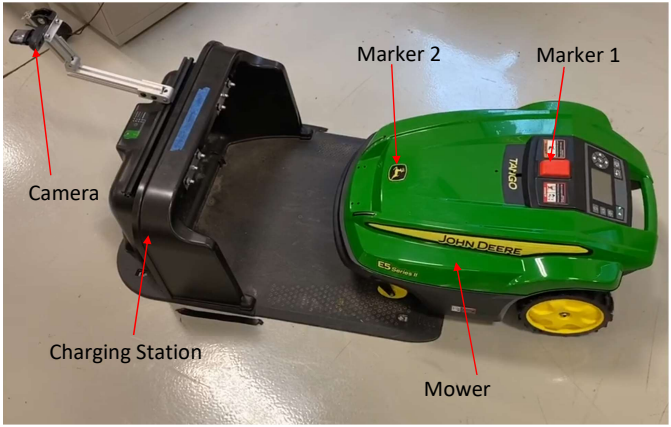}
\caption{Real-world docking scenario. The mower is shown approaching the charging station under the direction of the RL network, which receives inputs from the YOLO network which localizes the markers in the camera feed. Marker~1 is the stop button and marker~2 is the John Deere logo.}
\label{fig:env}
\end{figure}

In this paper, we have used a combination of a supervised and reinforcement learning algorithms to locally navigate the robot mower and orient it toward the docking station. Firstly, using transfer learning, we train a YOLO network to detect two pre-existing markers on the robot mower to provide positioning and orientation information. Secondly, using the output of the object detection network, we train a DQN agent to learn how to move the robot mower to the desired position employing a curriculum training technique. The real-world scenario is shown in Fig. \ref{fig:env}.

\section{Background} 
\subsection{YOLO}
Real time object detection has been a significant achievement of deep convolutions neural networks (CNN) in the field of computer vision. Using residual connections, deep CNNs can extract complex features from the observed image and are highly accurate in localizing different objects \cite{he2016deep}. Generally, these networks are trained on multiple object training sets. Given an image, they draw a bounding box, labeled with the object's tag, around it. The R-CNN algorithm \cite{girshick2014rich} is arguably the pioneer object localization algorithm and, ever since its introduction, many other algorithms have been proposed for the purpose of object detection, including modified versions of R-CNN such as fast R-CNN and faster R-CNN \cite{girshick2015fast, ren2016faster}. One of the fastest and most accurate object detection algorithms is You Only Look Once (YOLO), which achieves object detection using a fixed-grid regression \cite{redmon2016you}.

\subsection{Deep Q-Learning}
Value learning is a way of approaching reinforcement learning problems. Deep Q-Learning (DQN) is a value-learning algorithm that has demonstrated success on a wide range of tasks, including achieving human-level control in Atari games~\cite{mnih2015human}. The algorithm combines the traditional Q-learning update with neural network function approximation. Similarly to many RL algorithms, the problem is modeled as a discrete-time Markov Decision Process (MDP). At any time, the agent is in a certain state of the environment's state space, $S = \{s_{1}, s_{1}, ..., s_{n}\}$ and has some corresponding actions available from environment's action space, $A = \{a_{1}, a_{1}, ..., a_{m}\}$, which influence the transition to the next state. Transitioning from one state to another provides the agent with a reward $r$, and the goal of agent is to maximize the sum of its discounted rewards, $\sum_{t=0}^{\infty}\gamma^{t}r_{t}$, where $0<\gamma\leq1$ is the discount factor. The state-action value function is denoted by $Q(s,a)$, and maps a pair of a state and an action to a real number; $Q: S{\times}A \rightarrow \mathbb{R}$. The Q-learning update \cite{sutton2018reinforcement} is
\begin{align*}
    Q(s_{t},a_{t}) &\leftarrow Q(s_{t},a_{t}) \nonumber \\
    & \qquad + \alpha\Big(r_{t}-Q(s_{t},a_{t})+\gamma  \displaystyle\max_{a^\prime}Q(s_{t+1},a^\prime)\Big)
\end{align*}
where  $\alpha$ is the learning rate. Mnih \emph{et al.} \cite{mnih2015human} used the Q-learning algorithm in conjunction with deep neural networks and a replay buffer to produce the DQN (Deep Q-Network) algorithm. This uses two networks, the primary and target. The primary network is the one that is being updated using Stochastic Gradient Descent (SGD) at every iteration. The target network is the latest copy of the primary network, and it is used for evaluating the action values. The target network is updated once in every $N \in \mathbb{N}$ iterations to evaluate the action values using a recent version of the primary network. However, the max operator in DQN algorithm is prone to overestimating the state-action value function since it selects the maximum value of the same Q network. To mitigate this function approximation overestimation, one needs to decouple the selection and evaluation tasks. This leads to the Double DQN algorithm \cite{van2016deep}, which is used in this research:

\begin{algorithm}
\caption{Double DQN}
\begin{algorithmic}[1]
\State initialize the primary network $Q_{\theta}$, the target network $Q_{\theta{^\prime}}$, the replay buffer $D$, and $\tau\ll 1$
\For{each iteration}
\For{each environment step}
\State observe state $s_{t}$ and select $a_{t}\sim\pi(s_{t},a_{t})$
\State execute $a_{t}$ and observe the next state $s_{t+1}$
\State \qquad and reward $r_{t}=R(s_{t},a_{t})$
\State store $(s_{t}, a_{t}, r_{t}, s_{t+1})$ in the replay buffer $D$
\EndFor
\For{each update step}
\State sample $e_{t}=(s_{t}, a_{t}, r_{t}, s_{t+1})\sim D$
\State compute target $Q$ value:
\State \qquad $Q^{*}(s_{t},a_{t}) \approx r_{t} + \gamma Q_{\theta^{\prime}}(s_{t+1},\displaystyle\argmax_{a^{\prime}}Q_{{\theta}}(s_{t+1},a^{\prime}))$
\State perform gradient descent step on
\State \qquad $(Q^{*}(s_{t},a_{t})-Q_{\theta}(s_{t},a_{t}))^2$
\State update the target network parameters: $\theta^{\prime} \leftarrow \tau\theta+(1-\tau)\theta^{\prime}$
\EndFor
\EndFor
\end{algorithmic}
\end{algorithm}

\section{Simulation and Training}
\subsection{Motivation for Object Detection with RL}
The charging station of the mower is viewed by a Logitech c270 webcam, and the aim is to navigate the mower toward the docking station and either stop it at a desired position or to help it dock, using only the vision input. Therefore, the environment of the problem we are trying to solve, the video feed of the camera, is not only very high dimensional, but it is also dependent on where the setup is located, and it varies from yard to yard. Hence, we lower the dimensionality of the environment and extract key features of the video feed to both improve system robustness and accelerate RL training.

We use the YOLO algorithm to locate bounding boxes around two markers on the mower, one in the front and the other at the back of the top surface of the mower. The output of the YOLO network is then passed to the RL network as input. Accordingly, the RL agent's observation space of the world is low dimensional and it is easier to train in this space. The fact that the two markers are located at different ends of the mower allows the RL agent to sense the angle that the mower makes with the straight line toward the docking station. This is due to the fact that both bounding boxes will be in the center of the image if the the mower is exactly oriented toward the docking station. Therefore, in the setup designed in this paper, the centers of the bounding boxes in the picture are the information that is passed to the RL network, which outputs linear velocity and steering rate as actions.

\subsection{RL Simulation Environment}
\begin{figure}[b]
\centering
        \includegraphics[clip, trim=2.5cm 14cm 0.5cm 2.5cm, width = 0.55\textwidth]{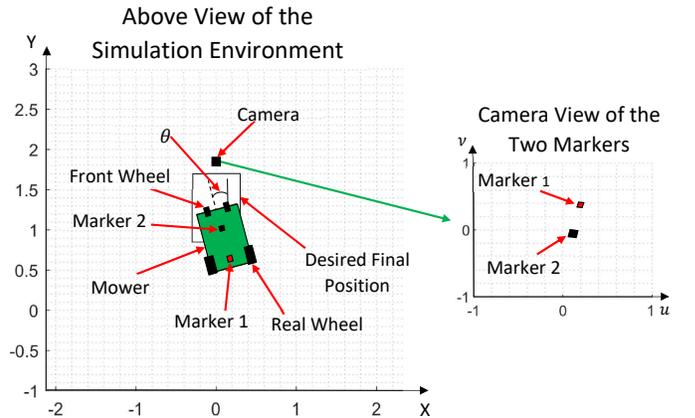}
\caption{Simulation environment. Left: the view from above of the mower approaching the charging station. Right: the simulated camera view of the two markers.}
\label{fig:main1}
\end{figure}

We have designed a simple simulation environment for training the RL agent. The kinematics of the agent in the simulation environment are ODEs driven by the linear velocity and angular steering rate. The simulation environment simulates the motion and computes the view of the markers from the point of view of the camera. The markers are drawn as quadrilaterals in the simulation, one in red and the other in black, and the rest of the simulated image is just a white background, representing the remainder of the environment. The mower initially starts at a uniformly distributed random position, $(X,Y) \sim U(-0.2,0.2)\times U(-0.2,0.2)$ in meters, and orientation, $\theta \sim U(-30,30)$ in degrees. Throughout the paper, $X$ and $Y$ coordinates correspond to the position of the rear axle of the mower. The environment is shown in Fig.~\ref{fig:main1}. We stop training the agent when the average reward reaches a certain amount, which was set experimentally. Moreover, in training, we manually reset the mower when its Y component exceeds 1{\rm\ m}. The mower stops when the $Y$ component of its position exceeds 1~m and the goal is to minimize the $X$ and $\theta$ offsets from zero when it stops. This target goal was chosen because it enables the two metallic rods on the front of the physical mower to connect to the metal pads in the charging station.

\begin{figure}
\centering
        \includegraphics[clip, trim=0cm 5cm 0cm 8.5cm, width = 0.5\textwidth]{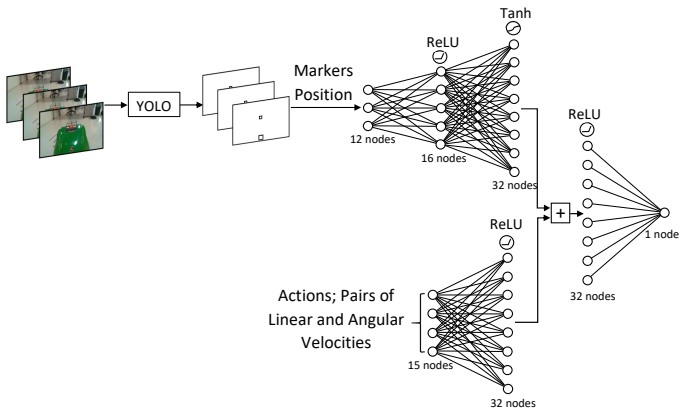}
\vspace{-3em}
\caption{The RL network architecture. The observation head receives labeled markers in the last three time-steps of the environment and feeds it through the network. The action head also passes each action through a single layer. The last layers of each head are added together and passed through the final layer to output the $Q$ values.}
\label{fig:network}
\end{figure}

The DQN agent takes the center positions of the bounding boxes at the last three time-steps as its input. The reason for using the data at the last three time-steps is to provide the agent with some information about the past to allow the estimation of velocity and acceleration. The agent outputs the desired linear velocity and angular rate based on its observation. The DQN network is shown in Fig.~\ref{fig:network}, where the state component has two fully connected layers with 16 and 32 neurons and the action component has a single layer with 32 neurons.

\begin{figure}
\centering
    \begin{subfigure}
        \centering
        {%
        \includegraphics[width = 0.21\textwidth]{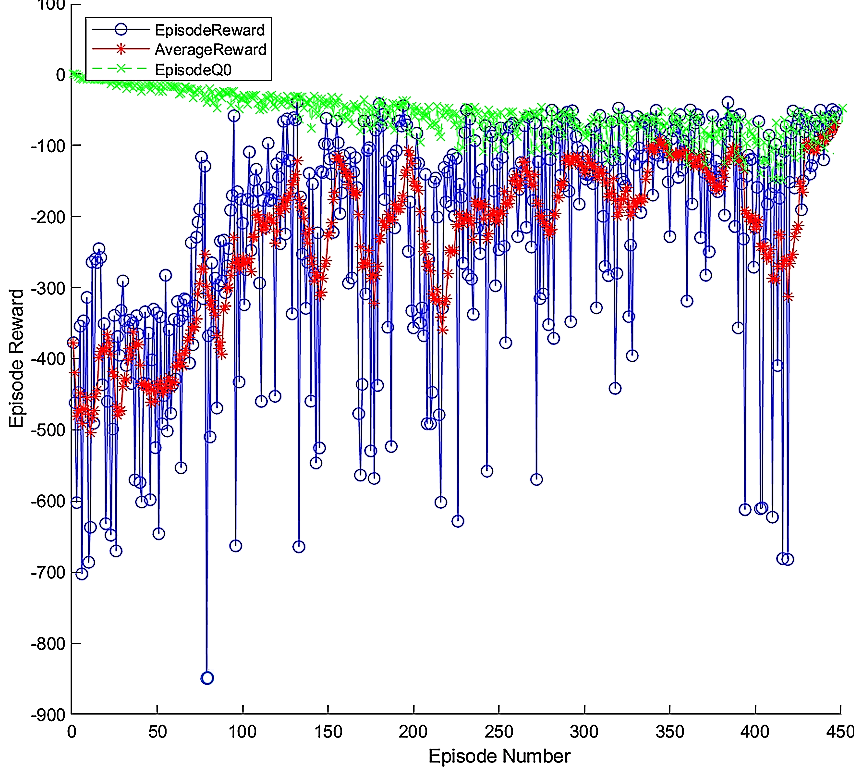}}
        \label{fig:t1}
    \hfill
    {%
        \includegraphics[width = 0.21\textwidth]{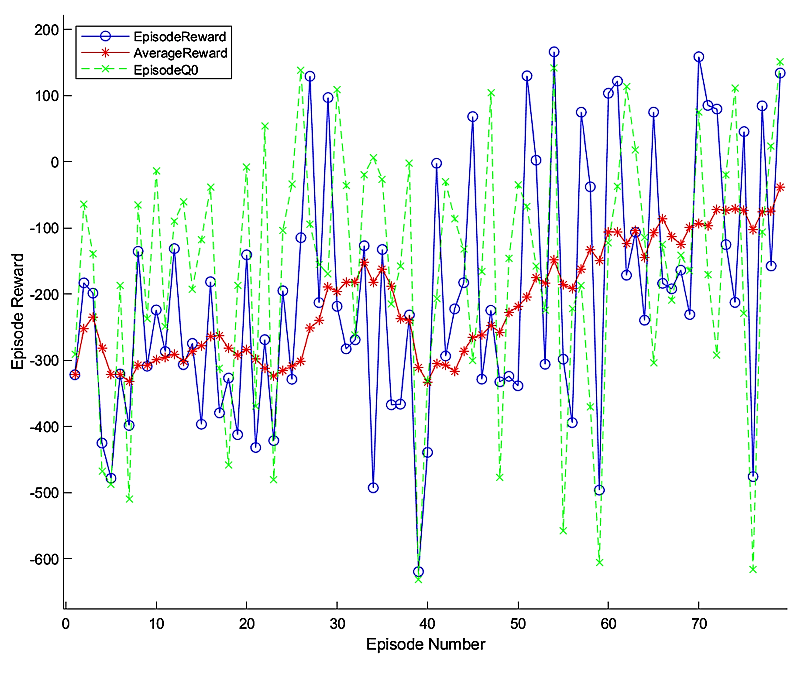}}
    \end{subfigure}
    \begin{subfigure}
        \centering
        {%
        \includegraphics[width = 0.21\textwidth]{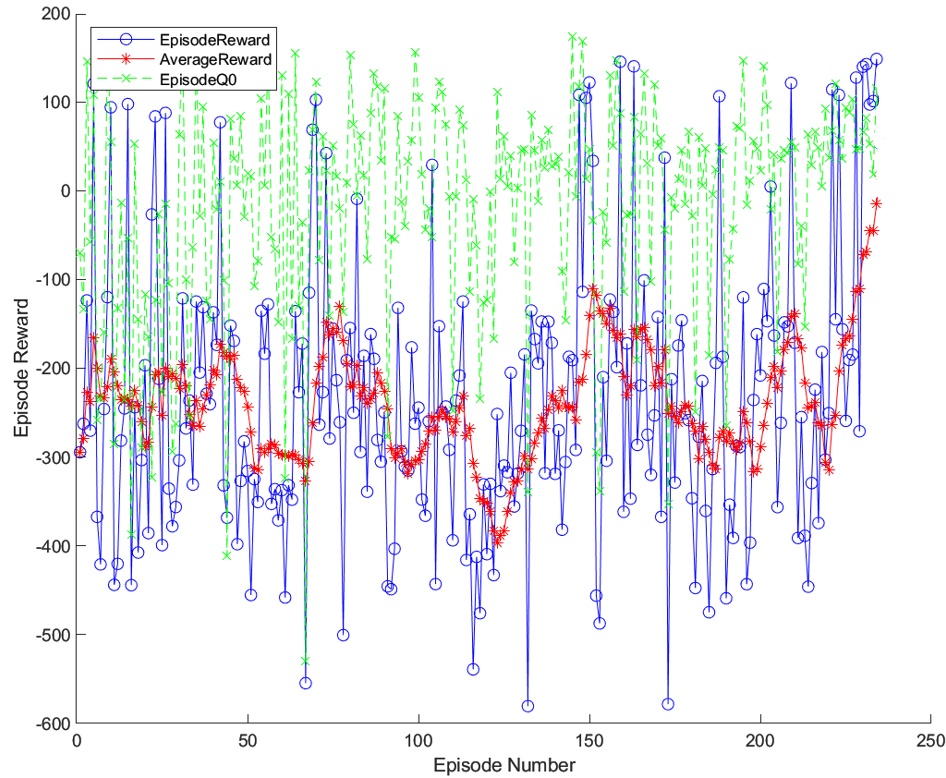}}
        \label{fig:t2}
    \hfill
    {%
        \includegraphics[width = 0.21\textwidth]{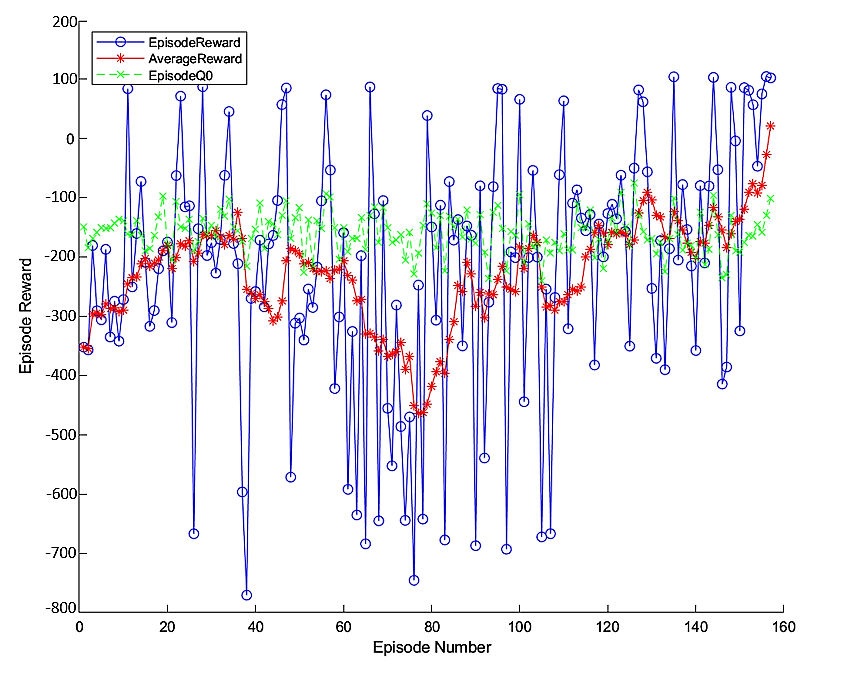}}
    \end{subfigure}

\caption{Training graphs of the agent during the four curriculum phases, with panels from top left, top right, bottom left, and bottom right corresponding to reward functions $r^1$ through $r^4$.}

\label{fig:train}
\end{figure}

For training the agent, we used curriculum training with four phases, numbered $i = 1,\ldots,4$. In the first phase the agent is rewarded to go forward, with a small reward for arriving near the target docking location. The subsequent phases increasingly reward more accurate docking. The four phases are distinguished by different reward functions $r^i_t$ defined by
\begin{equation*}
  r^{i}_{t} =
  \begin{cases}
  R^{i} & \text{if } Y \ge 1{\rm\ m} \text{ and } |u^{1}|,|u^{2}| < c^{i}_{1}, \\
  0 & \text{if } Y \ge 1{\rm\ m} \text{ and } |u^{1}|,|u^{2}| < 2c^{i}_{1}, \\
  -(10 |v^{1}(t) - v^{1}_{0}| + & \text{otherwise,}\\
  10|v^{2}(t) - v^{2}_{0}| + \\
  c^{i}_{2}|u^{1}| + c^{i}_{2}|u^{2}|)
  \end{cases}
\end{equation*}
where $(c^{i}_{1}, c^{i}_{2})$ are constants with values $(0.05,2)$, $(0.05,5)$, $(0.02,5)$, $(0.02,10)$ for $i = \{1,2,3,4\}$, respectively; $R^{i} = 0$ for $i = 1$, and $R^{i} = 150$ for $i = \{2,3,4\}$; $u^{1},v^{1},u^{2},v^{2} \in (-1,1)$ are the normalized coordinates of the center of the two markers; $v^{1}_{0},v^{2}_{0}$ are the $y$ components of the markers when the mower is at $Y=1{\rm\ m}$, $X=0{\rm\ m}$, $\theta=0^\circ$. The training data is shown in Fig.~\ref{fig:train}.

\begin{figure}
\centering
    \begin{subfigure}
       {%
        \includegraphics[clip, trim=0cm 20cm 0cm 0cm, width = 0.5\textwidth]{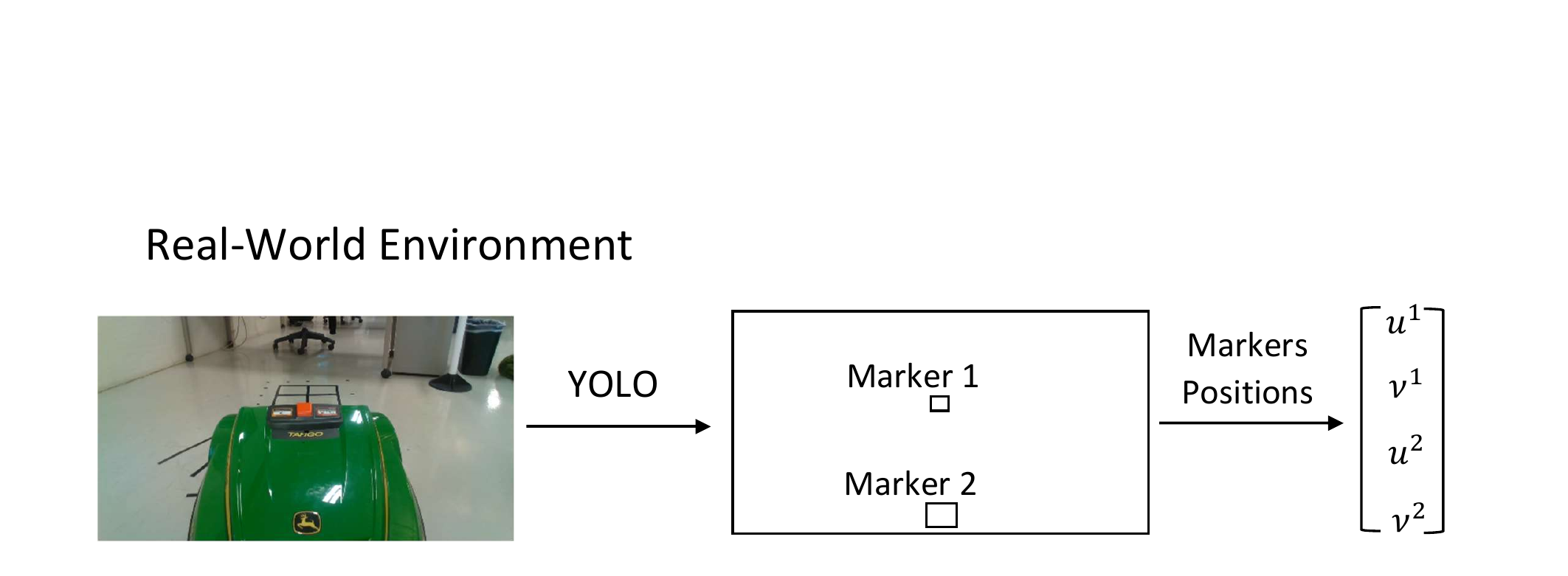}}
        \label{fig:t11}
        \hfill
    \end{subfigure}
    \begin{subfigure}
        {%
        \includegraphics[clip, trim=0cm 20cm 0cm 0cm, width = 0.5\textwidth]{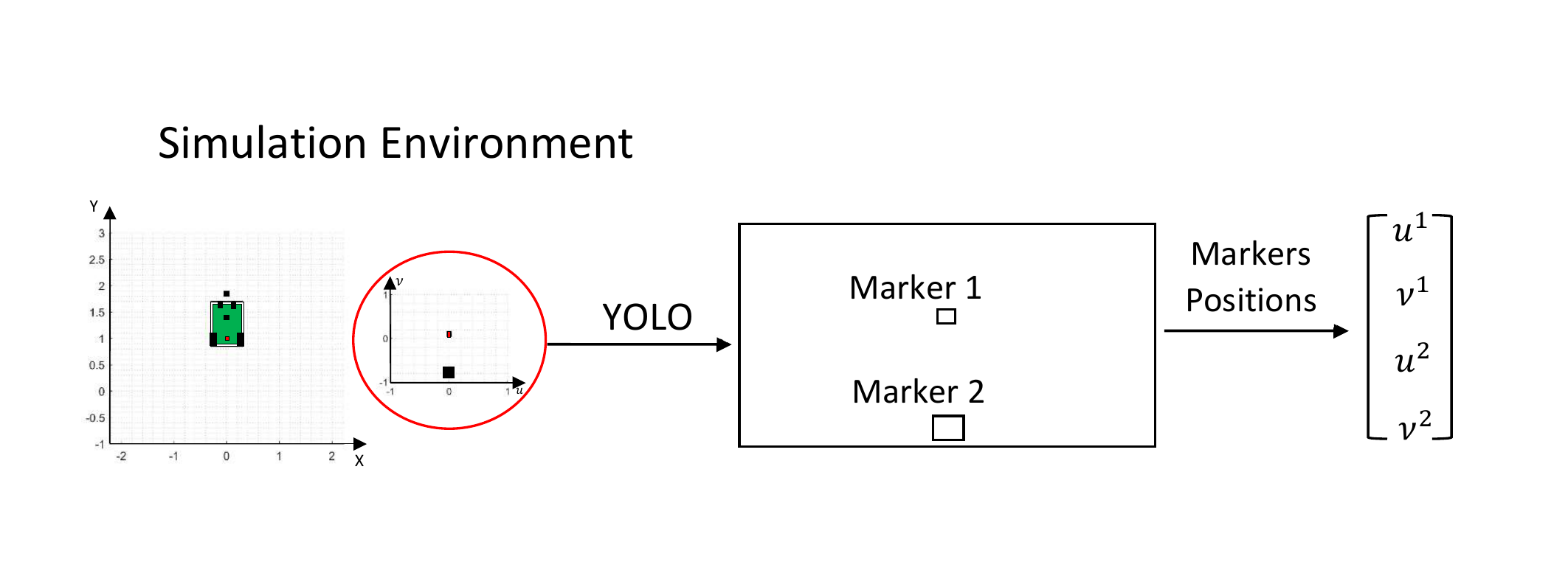}}
        \label{fig:t22}
        \hfill
    \end{subfigure}
\caption{The two YOLO networks. Top: YOLO network trained on real-wold mower data. Bottom: YOLO network trained on simulation images. In both cases the image is sent to the corresponding YOLO network and the outputs are the bounding boxes. Finally, the normalized positions of the center of the bounding boxes are extracted as a 4 dimensional vector.}
\label{fig:main2}
\end{figure}

\subsection{Object Detection Network}

There are two YOLO object detection networks used in this study, one for detecting the markers in the simulation and another for detecting the real-world markers from the actual camera images. For each of the networks, a pre-trained YOLO network was further trained to detect the markers. The training data for the networks is shown in Fig.~\ref{fig:main2}. Each of the networks was trained on about 3000 labelled images of the markers. The pre-trained feature extraction used ResNet50 \cite{he2016deep}, and the architecture is re-trained after the 20th ReLU function.

\section{Hardware Experiments} 
\subsection{Experiment Setup}
A Mosquitto MQTT broker was used to send the agent's actions to the mower via an Ethernet cable, and the connection between the computer and the mower was controlled by a Raspberry Pi. As in the simulation, the mower started from an initial position $(X,Y)$ and orientation $\theta$. The actions were sent to the mower at a frequency of $5$~Hz. The only input to the controlling RL agent was from the camera.

\subsection{Experiment Results}
\begin{figure}[ht]
\centering
\includegraphics[width = 0.4\textwidth, angle=0,origin=c,clip, trim=0.95cm 0cm 0cm 0cm]{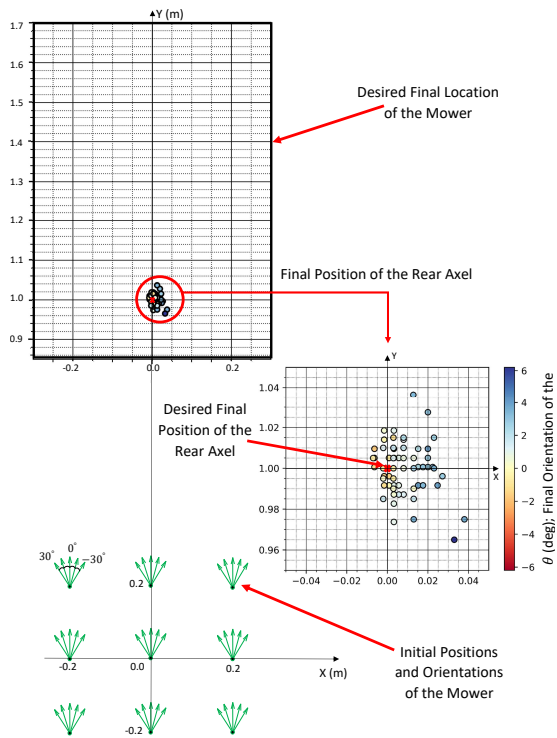}
\caption{Experiment configurations. The mower started from one of the positions and orientations shown in green, and from each initial configuration two tests were performed. The desired final position is marked by a red cross. The final positions of the mower are magnified and the final $X$, $Y$, and $\theta$ values are shown.}
\label{fig:realtest}
\end{figure}

A total of 90 tests were performed with the initial position and orientation of the mower $(X,Y,\theta)$ taken as all combinations of $X = -0.2,0,0.2$~m, $Y = -0.2,0,0.2$~m, and $\theta = -30,-15,0,15,30^\circ$, with each initial condition being tested twice. The performance of the agent was measured based the $X$, $Y$, and $\theta$ offsets of the mower when it stopped (when its observed $Y$ component has exceeded 1~m). The experiment data is shown in Fig.~\ref{fig:realtest} and summary results are given in Tab.~\ref{tab:testresults}, giving maximum absolute error, mean absolute error (MAE), and root mean squared error (RMSE). The mower had a maximum final position error of less than $4$~cm in both $X$ and $Y$ directions and a maximum final orientation error of less than $7^\circ$, representing successful positioning of the mower in all cases. The mean absolute error was less than 1~cm in both $X$ and $Y$ directions and less than $2^\circ$ in orientation.

\begin{table}[t]
\caption{Experiment results.}
\begin{center}
    \begin{tabular}{|c|c|c|c|}
    \hline
    \textbf{Error Measure}& \textbf{$X$ Offset (cm)}& \textbf{$Y$ Offset (cm)} & \textbf{$\theta$ Offset (deg)}\\ 
    \hline
    Max Abs. Error & 3.800 & 3.642 & 6.200\\
    \hline
    Mean Abs. Error & 0.822 & 0.934 & 1.533\\
    \hline
    RMSE & 0.896 & 1.182 & 1.661\\
    \hline
    \end{tabular}
    \label{tab:testresults}
\end{center}
\end{table}

\section{Conclusions}

We demonstrated a cheap and effective control system for autonomous docking of a robotic lawn mower (the John Deere Tango mower), using only vision from a single camera as the sensor. This system was shown to be robust in hardware tests, achieving centimeter-level docking precision.

The controller was a neural network trained using reinforcement learning (Double DQN) in a simple simulated environment. To avoid the need to simulate realistic vision inputs, we trained an object detection network (YOLO) to isolate two markers on the mower. The location of these markers was the only input to the controller agent, making it easy to produce similar inputs in simulation.

In addition to the ease of training, the use of an initial object detection network made the final system robust to different backgrounds and other environment variations. Our choice of markers was opportunistic (we used existing features on the mower) and it is likely that custom markers may be even better. We believe that this paradigm of an object detection network coupled with an RL agent could be an effective strategy for other robot motion control tasks.

\section*{Acknowledgment}
This research was supported by the John Deere Technology Innovation Center.

\bibliographystyle{IEEEtran}
\bibliography{References}
\end{document}